%% file: CameraReady_cvpr2022.tex
\documentclass[10pt,twocolumn,letterpaper]{article}

\usepackage{cvpr}              

\usepackage{graphicx}
\usepackage{threeparttable}
\usepackage{amsmath}
\usepackage{xcolor}
\usepackage{color, colortbl}
\usepackage{amssymb}
\usepackage{booktabs}
\usepackage{multirow}
\usepackage{enumitem}
\usepackage[accsupp]{axessibility}

\newcommand{\tablestyle}[2]{\setlength{\tabcolsep}{#1}\renewcommand{\arraystretch}{#2}\centering\footnotesize}
\newlength\savewidth
\newcommand{\myplus}[1]{\color{green}{\tiny{}}}
\newcommand{\myminus}[1]{\color{red}{\tiny{}}}

\usepackage[pagebackref,breaklinks,colorlinks]{hyperref}

\usepackage[capitalize]{cleveref}
\crefname{section}{Sec.}{Secs.}
\Crefname{section}{Section}{Sections}
\Crefname{table}{Table}{Tables}
\crefname{table}{Tab.}{Tabs.}

\def\confName{CVPR}
\def\confYear{2022}

\begin{document}


\title{TiG-BEV: Multi-view BEV 3D Object Detection via\\Target Inner-Geometry Learning}


\author{Peixiang Huang$^{1}$\footnotemark[2]\;,\;
Li Liu$^{2}$\footnotemark[2]\, \footnotemark[4]\;,\;
Renrui Zhang$^{3}$\footnotemark[2]\;,\;
Song Zhang$^{4}$\;,\;
Xinli Xu$^{2}$\;\;\\
Baichao Wang$^{2}$\;\;
Guoyi Liu$^{2}$\vspace{0.2cm}\\
$^1$Peking University\;
$^2$NIO\;
$^3$The Chinese University of Hong Kong\;\\
$^4$University of Chinese Academy of Sciences\\
{\tt\small liuli.ll9412@gmail.com,}\;
{\tt\small huangpx@stu.pku.edu.cn,}\;
{\tt\small 1155186671@link.cuhk.edu.hk,}\;\\
{\tt\small zhangsong20@mails.ucas.ac.cn,}\;
{\tt\small xxlbigbrother@gmail.com,}\;
{\tt\small \{baichao.wang,gary.liu\}@nio.com}\\
}
\maketitle

\input{cvpr_2022/tex/abstract}
\footnotetext[2]{Equal Contribution.}
\footnotetext[4]{Corresponding Author.}

\input{cvpr_2022/tex/intro}

\input{cvpr_2022/tex/relate}
\input{cvpr_2022/tex/method}

\input{cvpr_2022/tex/exp}

\section{Conclusion}
In this paper, we propose a novel target inner-geometry learning framework that enables the camera-based detector to inherit the effective foreground geometric semantics from the LiDAR modality. We first introduce an inner-depth supervision with target-adaptive depth reference to help the student learn better local geometric structures. Then, we conduct inner-feature distillation in BEV space for both channel-wise and keypoint-wise, which contributes to high-level inner-geometry semantics learning from the LiDAR modality. Extensive experiments are implemented to illustrate the significance of TiG-BEV for multi-view BEV 3D object detection. For future works, we will focus on exploring multi-modal learning strategy that can boost both camera and LiDAR modalities for unified real-world perception.
{\small
\bibliographystyle{ieee_fullname}
\bibliography{egbib}
}

\end{document}


\title{Supplementary Material:\\ Knowledge Distillation via the Target-aware Transformer}

\author{Sihao Lin$^{1,3}$\footnotemark[2]\;,\;
Hongwei Xie$^{2}$\footnotemark[2]\;,\;
Bing Wang$^{2}$,\;
Kaicheng Yu$^{2}$,\\
Xiaojun Chang$^{3}$\footnotemark[4]\;,\;
Xiaodan Liang$^{4}$,\;
Gang Wang$^{2}$\\
$^1$RMIT University\;
$^2$Alibaba Group\;
$^3$ReLER, AAII, UTS\;
$^4$Sun Yat-sen University\\
{\tt\small \{linsihao6, hongwei.xie.90, Kaicheng.yu.yt, xdliang328\}@gmail.com}\\
{\tt\small \{fengquan.wb, wg134231\}@alibaba-inc.com,}\;
{\tt\small xiaojun.chang@uts.edu.au}
}
\maketitle

\section{Asset Usage }
\label{sec:asset}
This work is built upon some public dataset and code assets. We appreciate their efforts. 
The benchmark dataset has been introduced in main paper. Here we list the URL, version, and license of the code assets that we used:

\begin{table*}[h]
    \centering
    \caption{Usage of Code assets.}
    \begin{tabular}{l|l|c|l}
        \toprule
         Exp.&\multicolumn{1}{c|}{URL} &Ver. &Licence \\\midrule
         ImageNet&{\tt \small https://github.com/yoshitomo-matsubara/torchdistill} &{\tt \small 7b883ec} &MIT\\ \midrule
         Cifar100&{\tt \small https://github.com/HobbitLong/RepDistiller}&{\tt \small 9b56e97} &BSD 2-Clause \\ \midrule
         \multirow{2}{*}{Pascal VOC}&{\tt \small https://github.com/jfzhang95/pytorch-deeplab-xception}&{\tt \small 9135e10} & MIT\\
         &{\tt \small https://github.com/clovaai/overhaul-distillation} &{\tt \small 76344a8} &MIT\\ \midrule
         \multirow{2}{*}{COCOStuff10k} &{\tt \small https://github.com/kazuto1011/deeplab-pytorch} &{\tt \small 4219467} &MIT\\
         &{\tt \small https://github.com/dvlab-research/ReviewKD} &{\tt \small cede6ea} &N/A\\
        \bottomrule
    \end{tabular}
    \label{tab:my_label}
\end{table*}


\footnotetext[4]{Corresponding Author.}
\footnotetext[2]{Equal contribution.}

\section{Additional Experiments}
\subsection{Comparison on COCOStuff10k}
For the experiments of semantic segmentation, we have compared our method to a variety of stat-of-the-art methods in the Section 4 of the main paper. In terms of COCOStuff10k, since some methods do not support this dataset, we re-implement them and the result is presented on Table~\ref{tab:supp_seg}. We found that our method is competitive and it outperforms the comparison methods.

\begin{table*}[h]
    \centering
    \caption{Comparison (mIoU\%) on COCOStuff10k.}
    \begin{tabular}{l|ccc}
    \toprule
         &ICKD~\cite{Liu2021ICKD} &Overhaul~\cite{Heo2019ACO} &Ours  \\ \midrule
        ResNet18 &27.22 &27.86 &28.75 \\   \midrule
        MobileNetV2 &26.64 &26.96 &28.05 \\
    \bottomrule
    \end{tabular}
    \label{tab:supp_seg}
\end{table*}


\subsection{Hyperparameters on Cifar-100}
We used Bayesian optimization to obtain the weight factors $\alpha$ and $\epsilon$ in Eq. 9. Here we show the searching result on different backbones (See Table~\ref{tab:supp_cifar100_p}). We found that in most cases (4 out of 6), $\epsilon$ is greater than $\alpha$, which indicates that our proposed objective is more important than the standard Cross-entropy during distillation. For instance, in the distillation VGG13$\rightarrow$VGG8, $\epsilon$ is 8 and $\alpha$ is only 0.1. We also found that for the similar architectures, the searching result is similar, \eg, when WRN-40-2 and ResNet110 are selected as teacher.

\begin{table*}[]
    \centering
    \caption{Coefficients $\alpha$ and $\epsilon$ on different backbones on Cifar-100.}
    \vspace{0mm}
    \begin{tabular}{c|ccccccc}
    \toprule
        Teacher & WRN-40-2 & WRN-40-2  & ResNet56 & ResNet110 & ResNet110 & ResNet32$\times$4 & VGG13 \\
        Student & WRN-16-2 & WRN-40-1  & ResNet20 & ResNet20  & ResNet32  & ResNet8$\times$4  & VGG8 \\ \midrule 
        $\alpha$   &0.8 &0.7 &0.8 &1 &1 &6 &0.1\\
        $\epsilon$ &4 &3.6 &0.4 &0.75 &1 &39 &8\\
        \bottomrule
         
    \end{tabular}
    \label{tab:supp_cifar100_p}
\end{table*}

\begin{table*}[t]
\centering
\vspace{0mm}
\caption{Adding $\mathcal{L}_{\rm{KL}}$ on Cifar100.}
\vspace{0mm}
\resizebox{1.0\columnwidth}{!}{
\tablestyle{20pt}{1.0}
\begin{tabular}{l|cccc} 
\toprule
Teacher& WRN-40-2 &ResNet110 & ResNet32$\times$4 & VGG13 \\
Student& WRN-16-2 &ResNet20  & ResNet8$\times$4  & VGG8 \\ \midrule
KD &74.92 &70.67 &73.33 &72.98 \\
FitNet+KD &75.12 &70.67 &74.66 &73.22 \\
AT+KD &75.32 &70.97 &74.53 &73.48 \\
SP+KD &74.98 &71.02 &74.02 &73.49 \\
CC+KD &75.09 &70.88 &74.21 &73.04 \\
RKD+KD &74.89 &70.77 &73.79 &72.97 \\
PKT+KD &75.33 &70.72 &74.23 &73.25 \\
NST+KD &74.67 &71.01 &74.28 &73.33 \\
CRD+KD &75.64 &71.56 &75.46 &74.29 \\
ICKD+KD &75.57 &71.91 &75.48 &73.88 \\
Ours+KD &\textbf{76.08}&\textbf{72.16}&\textbf{75.54}&\textbf{74.35}\\
\bottomrule
\end{tabular}
}

\label{tab:cifar_add_kl}
\vspace{0mm}
\end{table*}

\subsection{Adding KD loss on Cifar-100}
We report the result of our method in Table~\ref{tab:cifar_add_kl} with $\mathcal{L}_{\rm{KL}}$ loss to compare with the baselines under the same settings. Our method with KD loss surpasses all the baselines again. 

\subsection{Feature Visualization}
We further visualize the feature map and the associated TaT map to intuitively understand the functionality behind the proposed Target-aware Transformer. As exhibited in Figure~\ref{fig:supp_vis},  we visualize the feature maps of student before and after distillation, which are compared to the feature map of teacher. The teacher backbone is ResNet34 and student backbone is ResNet18. The input images are randomly selected from ImageNet validation set. While the 4-th block (\ie distillation layer) of ResNet34 and ResNet18 has 512 channels, we visualize 64 channels for better visualization. 

Obviously, the reconfigured student feature (3rd column) has a more similar pattern with teacher feature (4th column), which demonstrates that TaT can effectively adapt the student to mimic the teacher. In terms of the TaT map, which controls the intensity of semantic aggregation, it is close to the identity matrix. Recall that we apply the linear function $\phi(\cdot)$ on student feature $f^s$.
And the TaT map will be further applied on $\phi(f^{s})$ to reconfigure the student feature, which is lately asked to minimize the L$_2$ distance with teacher feature. When the TaT map is an identity matrix, it means that $\phi(f^{s})$ can reconstruct the teacher feature on its own. However, since TaT map is not strictly the identity matrix, it indicates that each pixel of $\phi(f^{s})$ still needs to \textit{borrow} the semantic from other position (mostly neighborhood) to enhance itself. Indeed, by aggregating the semantic from neighbors, each pixel increases the receptive field and thus semantic capacity. This demonstrates the semantic mismatch between student and teacher due to the variation on network depth and width.

\begin{figure*}[t]
    \centering
    \includegraphics[scale=1]{cvpr_2022/supp-2.pdf}
    \vspace{-1mm}
    \caption{
    \textbf{Visualization of feature map and TaT map.} The input is selected from ImageNet validation set. The teacher backbone is ResNet34 and student backbone is ResNet18. The feature map of the distillation layer (4-th block) has been visualized. While there are 512 feature channels in total, we visualize 64 channels for better visualization. Through the Target-aware transformer, we found that the reconfigured student feature (3rd column) has a similar pattern with teacher feature (4th column). 
    The associated TaT map has also been visualized, which indicates the student would aggregate the semantic mostly from neighbor to enhance its pixels. 
    }
    \label{fig:supp_vis}
    \vspace{-4mm}
\end{figure*}
{\footnotesize
\bibliographystyle{ieee_fullname}
\bibliography{egbib}
}

%% file: cvpr_2022/tex/abstract.tex
\begin{abstract}

To achieve accurate multi-view 3D object detection, existing methods propose to benefit camera-based detectors with spatial cues provided by the LiDAR modality, e.g., dense depth supervision and bird-eye-view (BEV) feature distillation. However, they directly conduct point-to-point mimicking from LiDAR to camera, which neglects the inner-geometry of foreground targets and suffers from the modal gap between 2D-3D features. In this paper, we propose the learning scheme of \textbf{T}arget \textbf{I}nner-\textbf{G}eometry from the LiDAR modality into camera-based BEV detectors for both dense depth and BEV features, termed as \textbf{TiG-BEV}. First, we introduce an inner-depth supervision module to learn the low-level relative depth relations between different foreground pixels. This enables the camera-based detector to better understand the object-wise spatial structures. Second, we design an inner-feature BEV distillation module to imitate the high-level semantics of different keypoints within foreground targets. To further alleviate the BEV feature gap between two modalities, we adopt both inter-channel and inter-keypoint distillation for feature-similarity modeling. With our target inner-geometry learning, TiG-BEV can effectively boost student models by different margins on nuScenes val set, e.g., \textbf{+2.3\%} NDS and \textbf{+2.4\%} mAP for BEVDepth, along with \textbf{+9.1\%} NDS and \textbf{+10.3\%} mAP for BEVDet without CBGS. Code will be available at \url{https://github.com/ADLab3Ds/TiG-BEV}.



\end{abstract}

%% file: cvpr_2022/tex/intro.tex
\vspace{-5mm}
\section{Introduction}
\label{sec:intro}

\begin{figure}[!t]
    \includegraphics[scale=0.062]{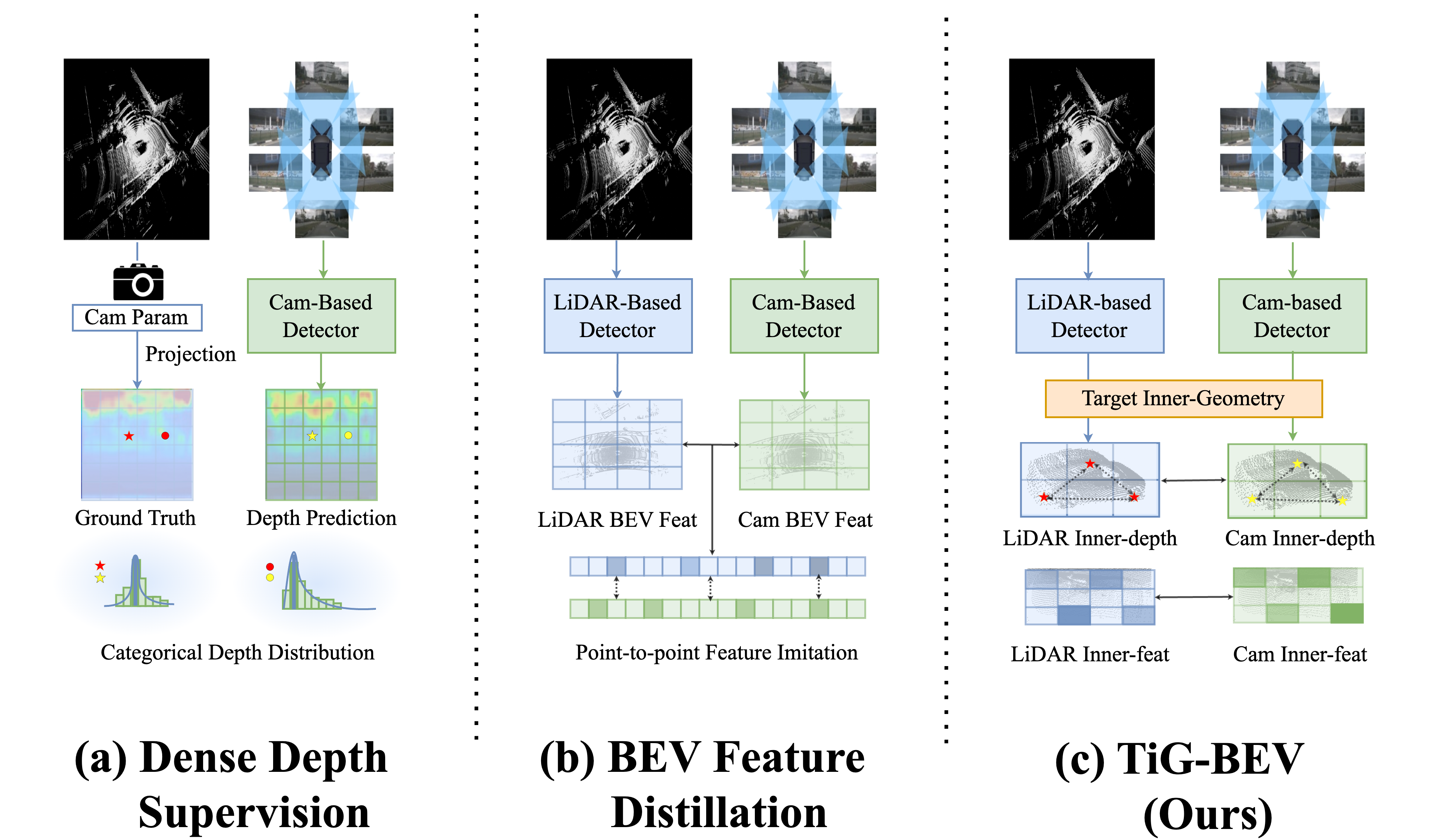}
    \caption{\textbf{Different LiDAR-to-Camera Learning Schemes:} (a) Dense Depth Supervision~\cite{b7,b51}, (b) BEV Feature Distillation~\cite{b9,b52}, (c) Our Target Inner-Geometry Learning (TiG-BEV).
    }
    \label{fig:teaser_figure}
\end{figure}

3D object detection aims to recognize and localize objects in 3D space, which is more challenging than its 2D counterpart and has achieved outstanding progress in various applications, such as robotics \cite{b1}, virtual reality \cite{b2}, and autonomous driving \cite{b3,b4,b5,b6}. Mainstream methods for 3D object detection can be categorized into LiDAR-based detectors \cite{b14, b3, b15} and camera-based detectors \cite{b7, b11, b12, b13, b47,b48}. Therein, LiDAR-based methods have attained excellent performance by taking 3D point clouds as input, which inherently contains sufficient spatial structures for accurate object localization. In contrast, camera-based methods are relatively low-cost with colored context information, but are constrained by the lack of geometric depth cues.

Considering the performance gap between camera-based and LiDAR-based detectors, 
existing methods leverage the spatial cues provided by the LiDAR modality to improve the accuracy of camera-based detectors, which are mainly in two schemes shown in Figure~\ref{fig:teaser_figure}. (1) Dense Depth Supervision (Figure~\ref{fig:teaser_figure} (a)), e.g., CaDDN~\cite{b51} and BEVDepth~\cite{b7}. These methods first project the input LiDAR points onto image planes using intrinsic and extrinsic camera parameters. Then, these derived depth maps are applied to explicitly supervise the categorical depth prediction within both foreground and background areas. (2) BEV Feature Distillation (Figure~\ref{fig:teaser_figure} (b)), e.g., CMKD~\cite{b52} and BEVDistill \cite{b9}. There methods adopt the teacher-student paradigm for BEV feature distillation. They force the BEV representation generated by the camera-based detector (student) to imitate that produced by a pre-trained LiDAR-based detector (teacher). By directly mimicking the BEV features, the student is expected to inherit the encoded high-level BEV semantics from the teacher.

However, existing methods fail to capture the inner-geometric characteristics of different foreground targets, i.e., the spatial contours and part-wise semantic relations. As examples, BEVDepth simply adopts pixel-by-pixel depth supervision without specializing the relative depth within objects, and BEVDistill applies foreground-guided distillation but neglects the inner relations of BEV features. In addition, methods~\cite{b9,b52} for BEV feature distillation directly enforce the channel-by-channel alignment between cross-modal BEV representations. Such strict mimicking might adversely affect the performance due to the modal gap between camera and LiDAR BEV features, i.e., visual appearances vs. spatial geometries.

\begin{figure}[!t]
\vspace{0.1cm}
    \centering
    \includegraphics[scale=0.12]{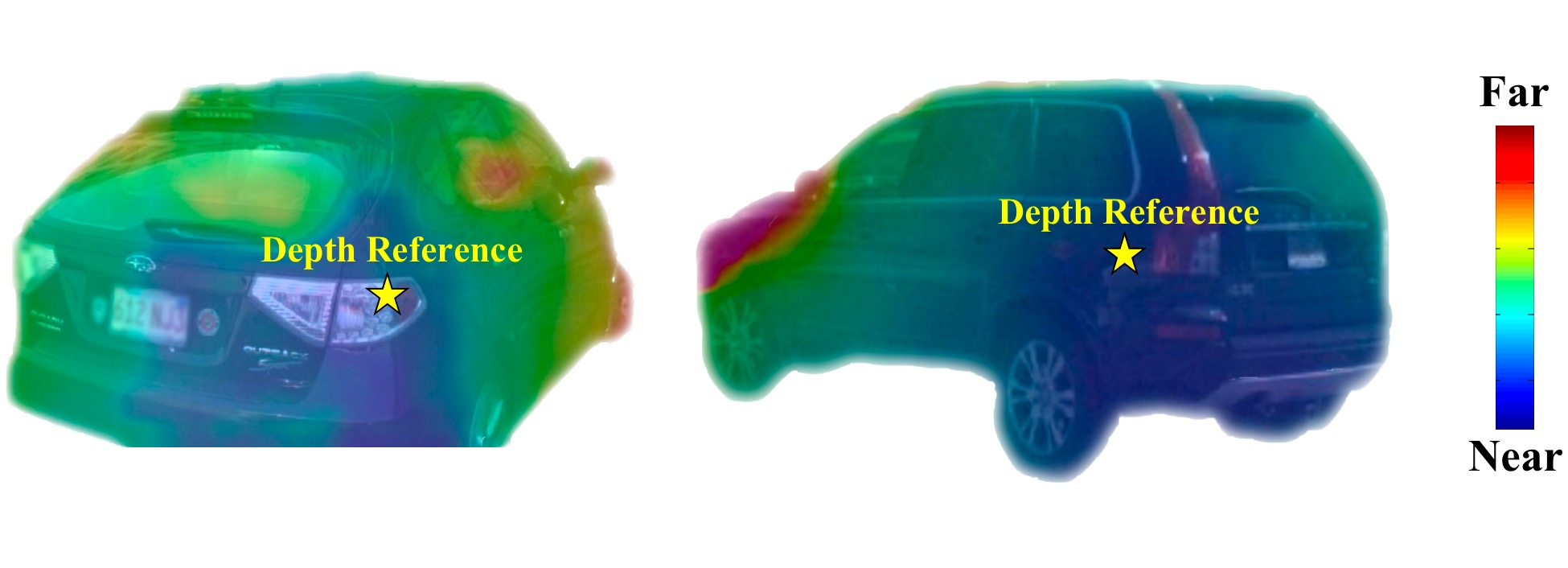}
    \caption{\textbf{Inner-depth Supervision.} We guide the camera-based detector to learn the relative spatial structures within the target foreground areas. A depth reference point (dotted in yellow) is adaptively selected to calculate relative depth values.}
    \label{fig:fig2}
\end{figure}

\begin{figure}[!t]
    \centering
    \includegraphics[scale=0.115]{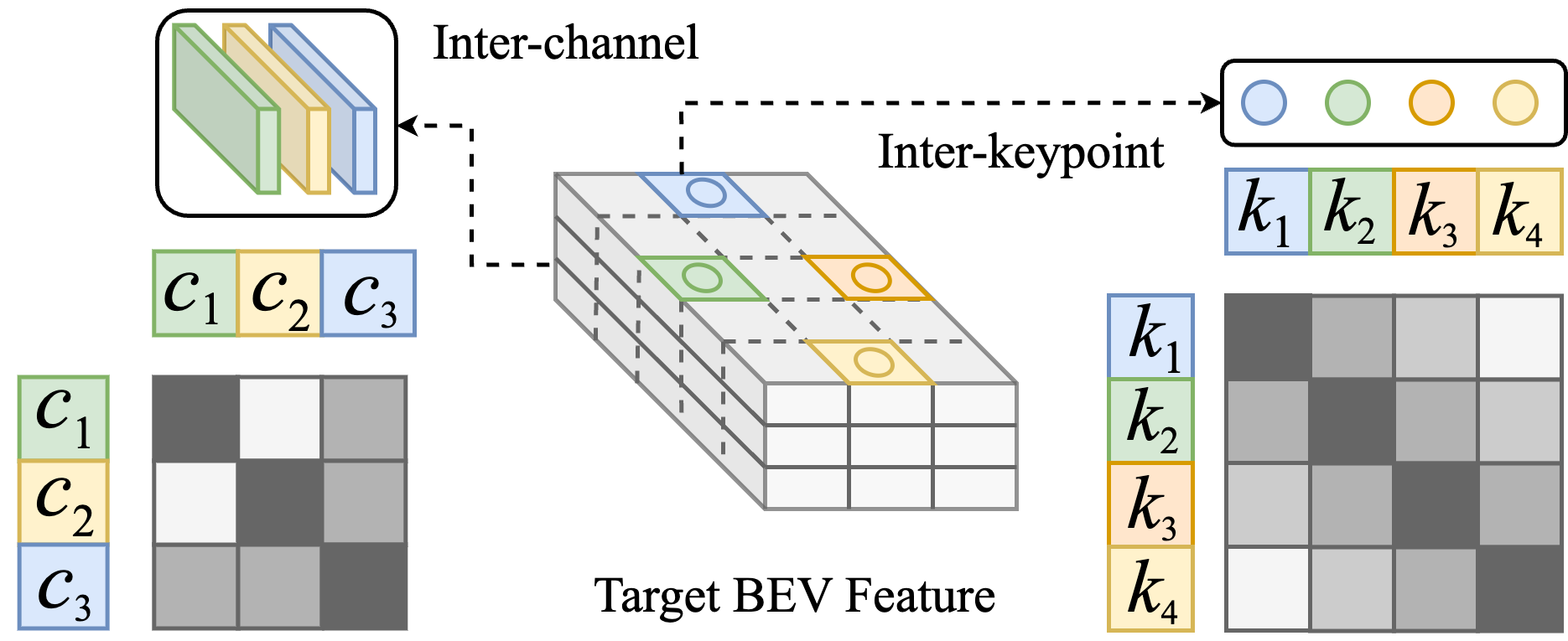}
    \caption{\textbf{Inner-feature BEV Distillation.} We respectively conduct inter-channel and inter-keypoint feature distillation in BEV space for the camera-based detector, which alleviates the cross-modal semantic gap and boosts inner-geometry learning.}
    \label{fig:fig3}
\end{figure}

To alleviate this issue, we propose a novel LiDAR-to-camera learning scheme, \textbf{TiG-BEV}, which involves the inner-geometry of foreground targets into the camera-based detectors for multi-view BEV 3D object detection. As shown in Figure~\ref{fig:teaser_figure} (c), we conduct simultaneous target inner-geometry learning for dense depth prediction and BEV feature generation.
First, besides the previous absolute depth map prediction~\cite{b7,b51}, we introduce an inner-depth supervision module within pixels of different foreground targets. A reference depth point is adaptively selected for each target to obtain the relative depth relationships shown in Figure~\ref{fig:fig2}, which contributes to high-quality depth map prediction with better target structural understanding. Second, we propose an inner-feature BEV distillation module, which imitates the high-level foreground BEV semantics produced by a pre-trained LiDAR-based detector. Different from previous dense and strict feature distillation~\cite{b9,b52}, we sample several keypoints within each BEV foreground area and guide the camera-based detector to learn their inner feature-similarities shown in Figure~\ref{fig:fig3}, which are in both inter-channel and inter-keypoint manners. In this way, the camera-based detector can not only inherit the high-level part-wise LiDAR semantics, but also relieve the modal gap by avoiding the strict feature mimicking. Through extensive experiments, we observe consistent performance improvement brought by TiG-BEV upon the baseline models. On nuScenes~\cite{b6} val set with equivalent settings, the powerful BEVDepth~\cite{b7} can be boosted by \textbf{+2.3\%} NDS and \textbf{+2.4\%} mAP, and BEVDet~\cite{b19} without CBGS can be further enhanced by \textbf{+9.1\%} NDS and \textbf{+10.3\%} mAP for BEVDet without CBGS, which well demonstrates the significance of our TiG-BEV.

The contributions of TiG-BEV are summarized below:
\begin{itemize}[leftmargin=*,itemsep=0pt,topsep=0pt]
    \item We introduce an inner-depth supervision module to model the relative depth relations of foreground targets, which leads to better depth map prediction.
    \item We propose an inner-feature BEV distillation module to transfer the well-learned knowledge from LiDAR modality to camera-based BEV representations. 
    \item Extensive experiments have verified the effectiveness of our approach to enhance the capacity for multi-view BEV 3D object detection.
\end{itemize}

%% file: cvpr_2022/tex/relate.tex
\section{Related Works}
\paragraph{Camera-based 3D Object Detection.} Camera-based 3D object detection has been widely used for applications like autonomous driving since its low cost compared with LiDAR-based detectors. FCOS3D \cite{b12} first predicts the 3D attributes of objects through the features around 2D centers and PGD \cite{b17} utilizes the relational graphs to improve the depth estimation for 3D monocular object detection. Further, MonoDETR~\cite{b47} introduces DETR-like~\cite{b18} architectures without complex post-processing.
Recently, Bird’s-Eye-View~(BEV), as a unified representation of surrounding views same as LiDAR-based detector, has attracted much attention. DETR3D \cite{b13} follows the DETR \cite{b18} to adopt the 3D reference points in BEV space by using object queries. BEVDet \cite{b19} utilizes the Lift-Splat-Shoot~(LSS) operation \cite{b20} to transform 2D image features into 3D Ego-car coordinate to generate 3D BEV feature. PETR \cite{b21} obtain the 3D position-aware ability by 3D positional embedding. Inspired by the recently developed attention mechanism, BEVFormer \cite{b11} and PolarFormer \cite{b22} automates the camera-to-BEV process with learnable attention modules and queries a BEV feature according to its position in 3D space. To further improve the detection performance, the temporal information has been introduced in BEVDet4D \cite{b23} and PETRv2 \cite{b24}, which achieve significant performance enhancement. Moreover, BEVDepth \cite{b7} observes that accurate depth estimation is essential for BEV 3D object detection supervised by projected LiDAR points. MonoDETR-MV~\cite{b48} proposes a depth-guided transformer for multi-view geometric cues, but predicts only foreground depth map without dense depth supervision. As a LiDAR-to-camera learning scehme, our TiG-BEV leverages the pre-trained LiDAR-based detector to improve the performance of camera-based detectors for multi-view BEV 3D object detection.

\paragraph{Depth Estimation.} Depth estimation is a classical problem in computer vision. These method can be divided into single-view depth estimation and multi-view depth estimation. Single-view depth estimation is either regarded as a regression problem of a dense depth map or a classification problem of the depth distribution. \cite{b26, b27, b28, b29, b30} generally build an encoder-decoder architecture to regress the depth map from contextual features. Multi-view depth estimation methods usually construct a cost volume to regress disparities based on photometric consistency \cite{b31, b32, b33, b34, b35, b62}. For 3D object detection, previous methods~\cite{b61,b51,b52} also introduce additional networks for depth estimation to improve the localization accuracy in 3D space.
Notably, MonoDETR \cite{b47,b48} proposes to only predict the foreground depth maps instead of the dense depth values, but cannot leverage the advanced geometries provided by LiDAR modality. Different from them, our TiG-BEV conducts inner-depth supervision that captures local sptial structures of different foreground targets.

\paragraph{Knowledge Distillation.} Knowledge Distillation has shown very promising ability in transferring learned representation from the larger model (teacher) to the smaller one (student). Prior works \cite{b37, b38, b39, b40} have been proposed to help the student network learn the structural representation for better generalization ability. These methods generally utilize the correlation of the instances to describe the geometry, similarity, or dissimilarity in the feature space. The following methods extend the teacher-student paradigm to many vision task, demonstrating its effectiveness including action recognition \cite{b41}, video caption \cite{b42}, 3D representation learning~\cite{b59,b49,b50,b60}, object detection \cite{b43, b44} and semantic segmentation \cite{b45, b46}. However, only a few of works consider the multi-modal setting between different sensor sources. For 3D representation learning, I2P-MAE~\cite{b49} leverages masked autoencoders to distill 2D pre-trained knowledge into 3D transformers. UVTR \cite{b25} presents a simple approach by directly regularizing the voxel representations between the student and teacher models. BEVDistill \cite{b9} transfer knowledge from LiDAR feature to the cam feature by dense feature distillation and sparse instance distillation. Our TiG-BEV also follows such teacher-student paradigm and effectively distills knowledge from the LiDAR modality into the camera modality,

%% file: cvpr_2022/tex/method.tex
\section{Method}
\label{sec:method}


\begin{figure*}[t!]
  \centering
  \includegraphics[width=\textwidth]{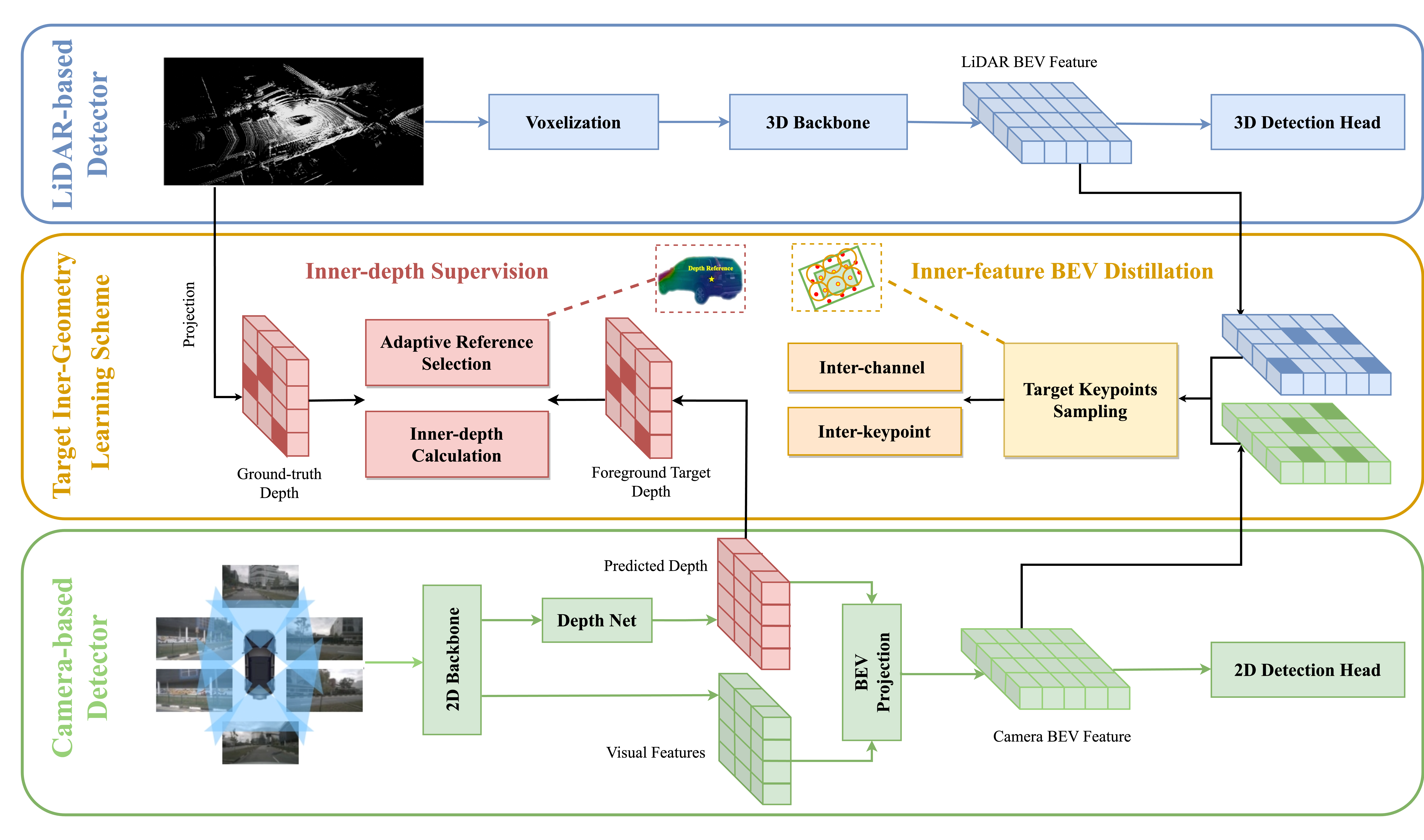}
  \caption{\textbf{Overall Framework of TiG-BEV,} which contains a pre-trained LiDAR-based detector as teacher, a camera-based detector as student, and a target inner-geometry scheme for cross-model learning. Our proposed learning paradigm effectively transfers the inner-geometry semantics of the LiDAR modality via two components, an inner-depth supervision (Section~\ref{sec:Inner-depth Supervision}) for foreground relative depth, and an inner-feature BEV distillation (Section~\ref{sec:Inner-feature BEV Distillation}) from both channel-wise and keypoint-wise.}
  \label{fig:framework}
\end{figure*}

The overall architecture of TiG-BEV is shown in Figure~\ref{fig:framework}, which consists of three components: the student camera-based detector, the teacher LiDAR-based detector, and our proposed target inner-geometry learning scheme. In Section~\ref{sec:Baseline Models}, we first introduce the adopted baseline models. Then, we specifically illustrate the designs of TiG-BEV for inner-depth supervision in Section~\ref{sec:Inner-depth Supervision} and inner-BEV feature distillation in Section~\ref{sec:Inner-feature BEV Distillation}. Finally in Section~\ref{sec:overall_loss}, we present the overall loss of our TiG-BEV for LiDAR-to-camera learning.

\subsection{Baseline Models}
\label{sec:Baseline Models}
\paragraph{Student Camera-based Detector.}
By default, we adopt BEVDepth~\cite{b7} as our student camera-based detector for multi-view 3D object detection. 
Given the input multi-view images (normally 6 views for a scene), the student model first utilizes a shared 2D backbone and FPN module~\cite{b54} to extract the $C$-channel visual features $\{F_i\}_{i=1}^6$, where $F_i \in {\mathbb{R}^{C\times H_v\times W_v}}$, and $H_v, W_v$ denote the size of feature maps. These features are fed into a shared depth network to generate the categorical depth map~\cite{b51}, $\{D_i\}_{i=1}^6$, where ${D_i}\in {\mathbb{R}^{D\times H_v\times W_v}}$, where D denotes the pre-defined number of depth bins. During training, BEVDepth adopts dense depth supervision for the predicted depth maps, which projects the paired LiDAR input onto multi-view image planes to construct pixel-by-pixel absolute depth ground truth, $\{D_i^{gt}\}_{i=1}^6$, where ${D_i^{gt}}\in {\mathbb{R}^{1\times H_v\times W_v}}$. 
Then, following~\cite{b20}, the multi-view visual features are projected into a unified BEV representation via the predicted depth maps, which is further encoded by a BEV encoder, denoted as $F^{2d}_{\rm bev}\in {\mathbb{R}^{C\times H_{\rm bev}\times W_{\rm bev}}}$. Finally, the detection heads are applied on top to predict objects in 3D space. We represent the two basic losses of the student model as $\mathcal{L}_{\rm{depth}}^{A}$ and $\mathcal{L}_{\rm{det}}$, respectively denoting the Binary Cross Entropy loss for dense absolute depth values and the 3D detection loss.

\paragraph{Teacher LiDAR-based Detector.}
We select the popular LiDAR detector CenterPoint~\cite{b53} as the teacher for target inner-geometry learning. Given the input point cloud data, CenterPoint voxelizes into grid-based data and utilizes a 3D backbone to obtain the $C$-channel LiDAR BEV feature $F^{3d}_{\rm bev}\in {\mathbb{R}^{C\times H_{\rm bev}\times W_{\rm bev}}}$, which has the same feature size as $F^{2d}_{\rm bev}$ from the student detector. As the CenterPoint has been well pre-trained, $F^{3d}_{bev}$ can provide the student BEV feature with sufficient geometric and semantic knowledge, espeically in the target foreground areas. Note that the LiDAR-based teacher is merely required during training for cross-modal learning, and for inference, only multi-view images are token as input for the camera-based detector.

\vspace{0.1cm}
\subsection{Inner-depth Supervision}
\label{sec:Inner-depth Supervision}

\begin{figure}[!t]
    \centering
    \includegraphics[scale=0.042]{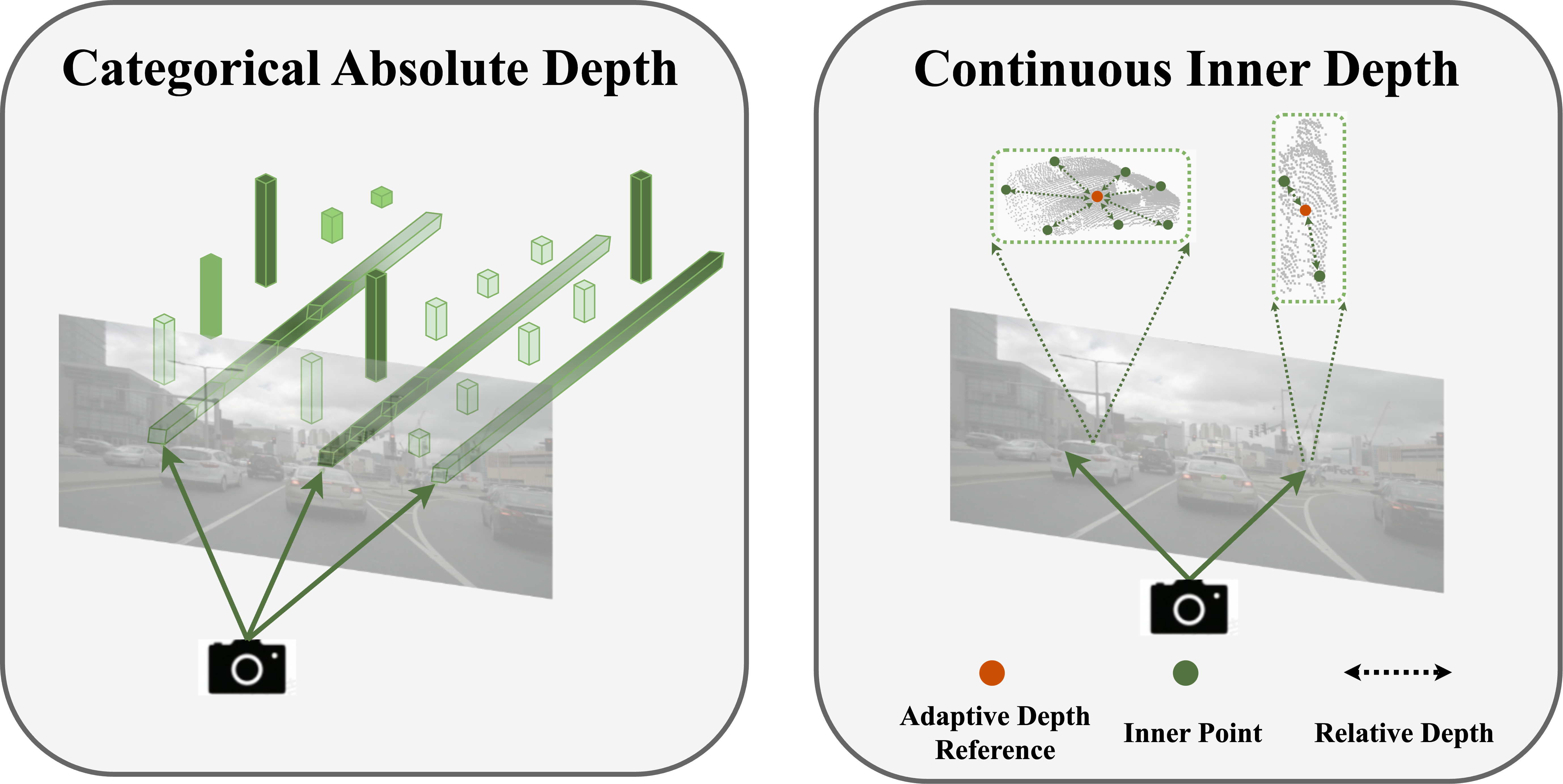}
    \caption{\textbf{Comparison of Categorical Absolute Depth and Continuous Inner Depth.} We adopt the inner-depth supervision with continuous depth values to guide the camera-based student to learn local spatial structures of foreground object targets.
    }
    \label{fig:relative_depth}
\end{figure}

In addition to the dense absolute depth supervision, we propose to guide the student model to learn the inner-depth geometries in different target foreground areas. As shown in Figure~\ref{fig:relative_depth} (a), for the instance level, the existing absolute depth supervision with categorical representation ignores the relative structural information inside each object and provide no explicit fine-grained depth signals. Therefore, we propose to additionally conduct inner-depth supervision with continuous values from the LiDAR projected depth maps shown in Figure~\ref{fig:relative_depth} (b), which effectively boosts the network to capture the inner-geometry of object targets.

\paragraph{Foreground Target Localization.}
To accurately obtain the inner-depth values, we first localize the foreground pixels for each object targets in the depth maps. Given the ground-truth 3D bounding boxes, we extract the corresponding 3D LiDAR points inside the box for each object target, and project them onto different image planes. In this way, we can attain the pixels within foreground object areas on both the predicted and ground-truth depth maps, $\{D_i, D_i^{gt}\}_{i=1}^6$. The foreground pixels can roughly depict the geometric contour of different target objects and well improve the subsequent inner-depth learning. We taking the $i$-th view as an example and omit the index $i$ in the following texts for simplicity. Suppose there exist $M$ target objects on the image, we denote the foreground depth-value set for the $M$ objects as $\{S_j, S_j^{gt}\}_{j=1}^M$, where each $\{S_j, S_j^{gt}\}$ includes the foreground categorical depth prediction and ground-truth depth values for the $j$-th target.

\paragraph{Continuous Depth Representation.}
Different from the categorical representation of absolute depth values, we represent the predicted inner depth of foreground targets by continuous values, which reflects more fine-grained geometric variations. For pixel $(x, y)$ of the $j$-th target object $S_j$, the predicted possibility of $k$-th depth bin is denoted as $S_j(x, y)[k]$, where $1\le k\le D$. Referring to MonoDETR~\cite{b47,b48}, we calculate the continuous depth value $d_j(x, y)$ for the pixel $(x, y)$ as
 \begin{equation}
 \label{eq:FM}
 \begin{aligned}
    d_j(x, y) = {\sum_{k=1}^D({d[k]\cdot S_j(x, y)[k]})},
 \end{aligned}
 \end{equation}
where $d[k]$ denotes the depth value of the $k$-th bin center. By this, we convert the categorical depth prediction of different target objects, $\{S_j\}_{j=1}^M$, into continuous representations, denoted as $\{\hat{S_j}\}_{j=1}^M$.

 \paragraph{Adaptive Depth Reference.}
 To calculate the relative depth values, we propose to utilize an adaptive depth reference for different foreground targets.
 Specifically, according to the predicted continuous depth values in $\{\hat{S_j}\}_{j=1}^M$, we select the pixel with the smallest depth prediction error as the reference point for each target, and correspondingly set its depth value as the depth reference, as shown in Figure~\ref{fig:relative_depth}. For the $j$-th target with the ground-truth inner-depth $\{\hat{S_j}, \hat{S^{gt}_j}\}_{j=1}^M$, we calculate the depth reference point $(x_r, y_r)$ by
  \begin{equation}
 \label{eq:FM}
 \begin{aligned}
    (x_r, y_r) = \mathop{\text{Argmin}}_{(x, y)\in \hat{S_j}} \left ({S^{gt}_j}(x, y) - {\hat{S_j}}(x, y)\right ).
 \end{aligned}
 \end{equation}
 Then, the predicted and ground-truth reference depth values are denoted as $d_j(x_r, y_r)$ and $d^{gt}_j(x_r, y_r)$, respectively. By adaptively selecting the reference point with the smallest error, the inner-depth distribution can dynamically adapt to objects with different shapes and appearances, which stabilizes the network learning for some truncated and occluded objects.

\paragraph{Inner-depth Calculation.}
On top of the reference depth value, we calculate the relative depth values within the foreground area of each target object. For pixel $(x, y)$ of the $j$-th target $\{\hat{S_j}, S_j^{gt}\}$, the predicted and ground-truth inner-depth values are formulated as
\begin{equation}
 \label{eq:FM}
 \begin{aligned}
    rd_j(x, y) &= d_j(x, y) - d_j(x_r, y_r),\\
    rd^{gt}_j(x, y) &= d^{gt}_j(x, y) - d^{gt}_j(x_r, y_r).
 \end{aligned}
 \end{equation}
 We denote the obtained relative depth-value sets for $M$ target objects as $\{\hat{R_j}, R_j^{gt}\}_{j=1}^M$. Finally, we supervise the inner-depth prediction of the student detector by an L2 loss, formulated as
 \begin{equation}
\label{eq:FM}
    \mathcal{L}_{\rm{depth}}^{R} = \sum_{j=1}^M ||\hat{R}_{j}-R^{gt}_{j}||_2.
\end{equation}

\subsection{Inner-feature BEV Distillation}
\label{sec:Inner-feature BEV Distillation}

\begin{figure}[!t]
    \centering
    \includegraphics[scale=0.17]{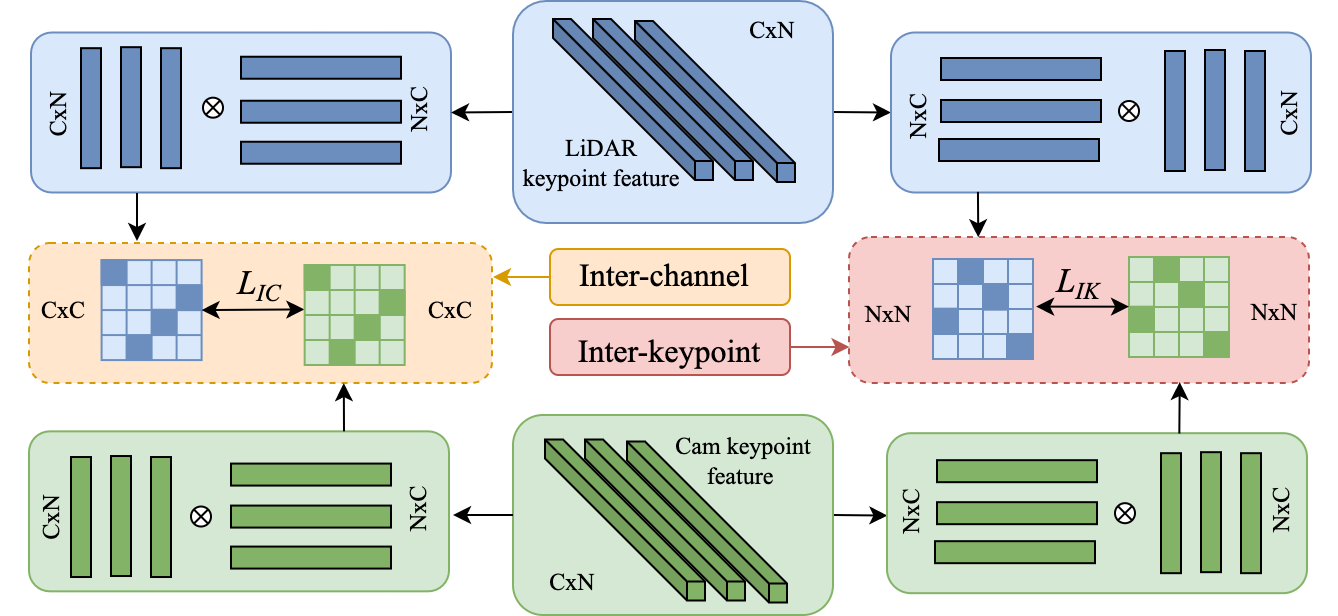}
    \caption{\textbf{Detials of Innter-feature BEV Distillation.} For each foreground area in BEV space, we represent rach target feature by a set of keypoints and conduct feature distillation in both inter-channel and inter-keypoint manners.
    }
    \label{fig:structure_attn}
\end{figure}

Besides the depth supervision for low-level spatial information, our TiG-BEV also adopts the inner-geometry learning for high-level BEV semantics from pre-trained LiDAR-based detectors. 
Previous works~\cite{b9,b52} for BEV distillation directly force the student to imitate the teacher's features point-to-point in the BEV space. In spite of the performance improvement, such strategies are constrained by the following two aspects. On the one hand, due to the sparsity of scanned point clouds, the LiDAR-based BEV features might contain redundant and noisy information in the background areas. Although BEVDistill~\cite{b9} utilizes foreground masks to alleviate this issue, such dense feature distillation still cannot provide focused and effective guidance to the student network. On the other hand, the camera-based and LiDAR-based BEV features depict different characteristics of the scene, respectively, visual appearances and spatial structures. Therefore, forcing the BEV features to be completely consistent between two modalities is sub-optimal considering the semantic gap. In our TiG-BEV, we propose an inner-feature BEV distillation (Figure~\ref{fig:structure_attn}) consisting of inter-channel and inter-keypoint learning schemes, which conducts attentive target features distillation and relieve the cross-modal semantic gap.


\paragraph{Target Keypoint Extraction.}
To distill the knowledge of LiDAR-based detectors only within sparse foreground regions, we extract the BEV area of each object target and represent it by a series of keypoint features.
Given the ground-truth 3D bounding box for each target, we first enlarge the box size for a little bit in the BEV space to cover the entire foreground area, e.g., object contours and edges. Then, we uniformly sample its BEV bounding box by $N$ keypoints, and adopt bilinear interpolation to obtain the keypoint features from the encoded BEV representations. From both camera-based $F^{2d}_{\rm bev}$ and LiDAR-based $F^{3d}_{\rm bev}$, we respectively extract the keypoint features for all $M$ object targets as $\{f_j^{2d}, f_j^{3d}\}_{j=1}^M$, where $f_j^{2d}, f_j^{3d} \in {\mathbb{R}^{N\times C}}$. By the uniform sampling, such BEV keypoints can well represent the part-wise features and the inner-geometry semantics of foreground targets.

\paragraph{Inter-channel BEV Distillation.}
Taking the $j$-th object target as an example, we first apply an inter-channel BEV distillation, which guides the student keypoint features to mimic the channel-wise relationships of the teacher's. Such inter-channel signals imply the overall geometric semantics of each object target. Compared with the previous channel-by-channel supervision, our inter-channel distillation can preserve the distinctive aspects of the two modalities, while effectively transfer the well pre-trained knowledge of LiDAR-based detectors. Specifically, we calculate the inter-channel similarities of both camera-based and LiDAR-based keypoint features, formulated as
\begin{equation}
    A_j^{2d} = f_j^{2d} {f_j^{2d}}^{\top};\ \ \ A_j^{3d} = f_j^{3d} {f_j^{3d}}^{\top},
\end{equation}
where $A_j^{2d}, A_j^{3d} \in {\mathbb{R}^{C\times C}}$ denote the feature relationships between different $C$ channels for the two modalities. For all $M$ objects in a scene, we adopt L2 loss between the two inter-channel similarities for feature distillation, formulated as
\begin{equation}
    \mathcal{L}_{\rm{bev}}^{{IC}}= \sum_{j=1}^{M} ||A_j^{3d}-A_j^{2d}||_2.
    \label{eq:fm}
\end{equation}

\input{cvpr_2022/tex/tables/nus_val_sota.tex}
\paragraph{Inter-keypoint BEV Distillation.}
\label{sec:anchor}
The inter-channel distillation guides the camera-based detector to learn the channel-wise diversity from the LiDAR-based teacher. However, it is conducted without considering the inner correlation of different keypoints within each object target, which is not capable of capturing the local geometries among different foreground parts, e.g., the front and rear of cars. To this end, we propose to utilize the inter-keypoint correlations of LiDAR-based BEV features and transfer such inner-geometry semantics into camera-based detectors. Analogous to the aforementioned inter-channel module, for the $j$-th target object, we calculate the inter-keypoint similarities in a transposed manner for the two modalities as
\begin{equation}
    B_j^{2d} = {f_j^{2d}}^{\top} {f_j^{2d}};\ \ \ B_j^{3d} = {f_j^{3d}}^{\top} {f_j^{3d}},
\end{equation}
where $B_j^{2d}, B_j^{3d} \in {\mathbb{R}^{N\times N}}$ denote the feature relationships between different $N$ keypoints respectively for camera and LiDAR. We also adopt L2 loss for all $M$ targets as
\begin{equation}
    \mathcal{L}_{\rm{bev}}^{{IK}}= \sum_{j=1}^{M} ||B_j^{3d}-B_j^{2d}||_2.
    \label{eq:fm}
\end{equation}
Then, the distillation loss for inter-channel and inter-keypoint features in BEV space is formulated as
\begin{equation}
    \mathcal{L}_{\rm{bev}}=
    \mathcal{L}_{\rm{bev}}^{{IC}}+
    \mathcal{L}_{\rm{bev}}^{{IK}},
    \label{eq:seg}
\end{equation}
where the two terms are orthogonal respectively for the channel-wise feature diversity and keypoint-wise semantic correlations.

\subsection{Overall Loss}
\label{sec:overall_loss}
To sum up, we benefit the student camera-based detector by target inner-geometry from two complementary aspects, i.e., an inner-depth supervision for low-level signals and an inner-feature BEV distillation for high-level semantics. They produce two losses as $\mathcal{L}^R_{\rm{depth}}$ and $\mathcal{L}_{\rm{bev}}$. Together with the original two losses, i.e., dense absolute depth supervision $\mathcal{L}^A_{\rm{depth}}$, and 3D detection $\mathcal{L}_{\rm{det}}$, the overall loss of our TiG-BEV is formulated as
\begin{equation}
    \mathcal{L}_{\rm{TiG}}=
    \mathcal{L}_{\rm{det}}+
    \mathcal{L}^A_{\rm{depth}}+
    \mathcal{L}^R_{\rm{depth}}+
    \mathcal{L}^{IC}_{\rm{bev}}+
    \mathcal{L}^{IK}_{\rm{bev}}.
    \label{eq:seg}
\end{equation}



%% file: cvpr_2022/tex/tables/nus_val_sota.tex
\begin{table*}[ht]
\centering
\caption{\textbf{Performance Comparison on nuScenes~\cite{b6} Val Set.} 'C' and 'L' denote the camera-based and LiDAR-based methods, which refer to the input data during inference. * denotes our implementation using BEVDet\cite{b19} codebase.
}
\resizebox{2\columnwidth}{!}{
\tablestyle{5pt}{1.2}
\begin{tabular}{c|c|c|c|cc|ccccc}
\toprule[1.2pt]
Method          & Modality & Backbone & Resolution & mAP↑  &NDS↑& mATE↓ & mASE↓ & mAOE↓ & mAVE↓ & mAAE↓   \\ \midrule
FCOS3D\cite{b12}          & C  &ResNet-101 & 900 $\times$ 1600 & 0.343& 0.415 & 0.725 & 0.263 & 0.422 & 1.292 & 0.153  \\
PGD\cite{b17}             & C  &ResNet-101 & 900 $\times$ 1600 & 0.369& 0.428 & 0.683 & 0.260 & 0.439 & 1.268 & 0.185  \\
MonoDETR\cite{b47}             & C  &ResNet-101 & 900 $\times$ 1600 & 0.372& 0.434 & 0.676 & 0.258 & 0.429 & 1.253 & 0.176  \\
DETR3D\cite{b13}          & C &ResNet-101  & 900 $\times$ 1600 & 0.303& 0.374 & 0.860 & 0.278 & 0.437 & 0.967 & 0.235  \\
PETR\cite{b21}            & C  &ResNet-101    & 512 $\times$ 1408 & 0.357 & 0.421& 0.710 & 0.270 & 0.490 & 0.885 & 0.224  \\
BEVFormer\cite{b11}       & C  &ResNet-101   & 900 $\times$ 1600 & 0.416 & 0.517& 0.673 & 0.274 & 0.372 & 0.394 & 0.198   \\
PETRv2\cite{b24}            & C  &ResNet-101    & 640 $\times$ 1600 & 0.421& 0.524 & 0.681 & 0.267 & 0.357 & 0.377 & 0.186  \\
MonoDETR-MV\cite{b48}            & C  &ResNet-101    & 640 $\times$ 1600 & 0.428 & 0.531 & 0.676 & 0.268 & 0.352 & 0.380 & 0.169  \\ \midrule
CenterPoint~\cite{b53} (Teacher)   & L &VoxelNet   & -          & 0.564 & 0.646& 0.299 & 0.254 & 0.330 & 0.286 & 0.191  \\ \midrule
BEVDet$^*$ \cite{b19} & C &ResNet-50   & 256 $\times$ 704 & 0.298& 0.379 & 0.725 & 0.279 & 0.589 & 0.860 & 0.245  \\ 
\rowcolor{gray!12} \textbf{+ TiG-BEV}     &C  &ResNet-50  & 256 $\times$ 704 & \textbf{0.331} & \textbf{0.411}& 0.678 & 0.271 & 0.589 & 0.784 & 0.218  \\ 
\rowcolor{gray!12}& &&& \textbf{\textcolor{blue}{+3.3$\%$} }& \textbf{\textcolor{blue}{+3.2$\%$}}&\textcolor{blue}{-4.7$\%$} & \textcolor{blue}{-0.8$\%$} & \textcolor{blue}{-0.0$\%$} & \textcolor{blue}{-7.6$\%$} & \textcolor{blue}{-2.7$\%$}   \\ 
\midrule
BEVDet4D$^*$ \cite{b23} & C &ResNet-50   & 256 $\times$ 704 & 0.322 & 0.451& 0.724& 0.277& 0.520 &0.366 &0.212  \\
\rowcolor{gray!12} \textbf{+ TiG-BEV}     & C &ResNet-50  & 256 $\times$ 704 & \textbf{0.356}& \textbf{0.477} & 0.648 & 0.273 & 0.517 & 0.364 & 0.210  \\ 
\rowcolor{gray!12}& &&& \textbf{\textcolor{blue}{+3.4$\%$} }& \textbf{\textcolor{blue}{+2.6$\%$}}&\textcolor{blue}{-7.6$\%$} & \textcolor{blue}{-0.4$\%$} & \textcolor{blue}{-0.3$\%$} & \textcolor{blue}{-0.2$\%$} & \textcolor{blue}{-0.2$\%$}   \\ 
\midrule
BEVDepth$^*$ \cite{b7} & C &ResNet-101   & 512 $\times$ 1408 & 0.416 & 0.521& 0.605 & 0.268 & 0.455 & 0.333 & 0.203  \\
\rowcolor{gray!12}   \textbf{+ TiG-BEV}   & C & ResNet-101  & 512 $\times$ 1408  & \textbf{0.440}& \textbf{0.544} & 0.570 & 0.267 & 0.392 & 0.331 & 0.201  \\ 
\rowcolor{gray!12}& &&& \textbf{\textcolor{blue}{+2.4$\%$} }& \textbf{\textcolor{blue}{+2.3$\%$}}&\textcolor{blue}{-3.5$\%$} & \textcolor{blue}{-0.1$\%$} & \textcolor{blue}{-6.3$\%$} & \textcolor{blue}{-0.2$\%$} & \textcolor{blue}{-0.2$\%$}   \\ 
\bottomrule[1.2pt]
\end{tabular}
}

\label{tab:nus_val_sota}
\end{table*}

%% file: cvpr_2022/tex/exp.tex
\vspace{0.2cm}
\section{Experiment}
\label{sec:experiment}
In this section, we first introduce our adopted dataset and implementation settings. Then, we conduct a series of experiments with detailed ablation studies to show the effectiveness of our approach.

\subsection{Experimental Settings}
\paragraph{Dataset and Metrics.}
We evaluate our TiG-BEV on nuScenes dataset\cite{b6}, which is one of the most popular large-scale outdoor public datasets for autonomous driving. It consists of 700, 150, 150 scenes for training, validation and testing, respectively. It provides synced data captured from a 32-beam LiDAR at 20Hz and six cameras covering 360-degree horizontally at 12Hz. We adopt the official evaluation toolbox provided by nuScenes, which reports the nuScenes Detection Score (NDS) and mean Average Precision (mAP), along with mean Average Translation Error (mATE), mean Average Scale Error (mASE), mean Average Orientation Error (mAOE), mean Average Velocity Error (mAVE), and mean Average Attribute Error (mAAE).

\begin{table}[t!]
\centering
\caption{\textbf{Performance Comparison without CBGS Strategy~\cite{b56}.} For all methods, we adopt ResNet-101 as the 2D backbone and $512\times 1408$ as the image resolution. * denotes our implementation.}
\resizebox{1\columnwidth}{!}
{\tablestyle{10pt}{1}
\begin{tabular}{c|cc}
\toprule[1.2pt]
          Method& mAP&NDS  \\
          \midrule
          BEVDet$^*$& 0.272 & 0.297 \\
          \rowcolor{gray!12}\textbf{+ TiG-BEV}& \textbf{0.375 (\textcolor{blue}{+10.3\%})} &\textbf{0.388 (\textcolor{blue}{+9.1\%})} \\\midrule
          BEVDet4D$^*$& 0.336 & 0.435 \\
          \rowcolor{gray!12}\textbf{+ TiG-BEV}& \textbf{0.409 (\textcolor{blue}{+7.3\%})} &\textbf{0.489 (\textcolor{blue}{+5.4\%})} \\\midrule
          BEVDepth$^*$& 0.393 & 0.487 \\ 
          \rowcolor{gray!12}\textbf{+ TiG-BEV}& \textbf{0.430 (\textcolor{blue}{+3.7\%})} &\textbf{0.514 (\textcolor{blue}{+2.7\%})} \\
    \bottomrule[1.2pt]
    \end{tabular}}
    \label{tab:aaaa}
\end{table}
\begin{table}[t!]
\centering
    \caption{\textbf{Comparison with BEVDistill~\cite{b9}.} $\dagger$ and * denote the implementation of BEVDistill and ours, respectively. We present the performance improvement of the learning methods correspondingly to their implemented baselines.}
    \resizebox{1\columnwidth}{!}
    {\tablestyle{10pt}{1.0}
    \begin{tabular}{c|cc}
    \toprule[1.2pt]
          Method&mAP&NDS  \\
          \midrule
          BEVDepth$\dagger$ &0.311       &  0.432     \\
          \rowcolor{gray!12}BEVDepth$^*$ &0.329       &  0.431     \\
          \midrule
          + BEVDistill &0.332 (\textcolor{blue}{+2.1\%})       &  0.454 (\textcolor{blue}{+2.2\%})     \\
          \rowcolor{gray!12}\textbf{+ TiG-BEV} &\textbf{0.366 (\textcolor{blue}{+3.7\%})}       &  \textbf{0.461 (\textcolor{blue}{+3.0\%})}     \\ 
    \bottomrule[1.2pt]
    \end{tabular}}
    \label{cccc}
\end{table}
\begin{figure*}[ht]
    \centering
    \includegraphics[scale=0.85]{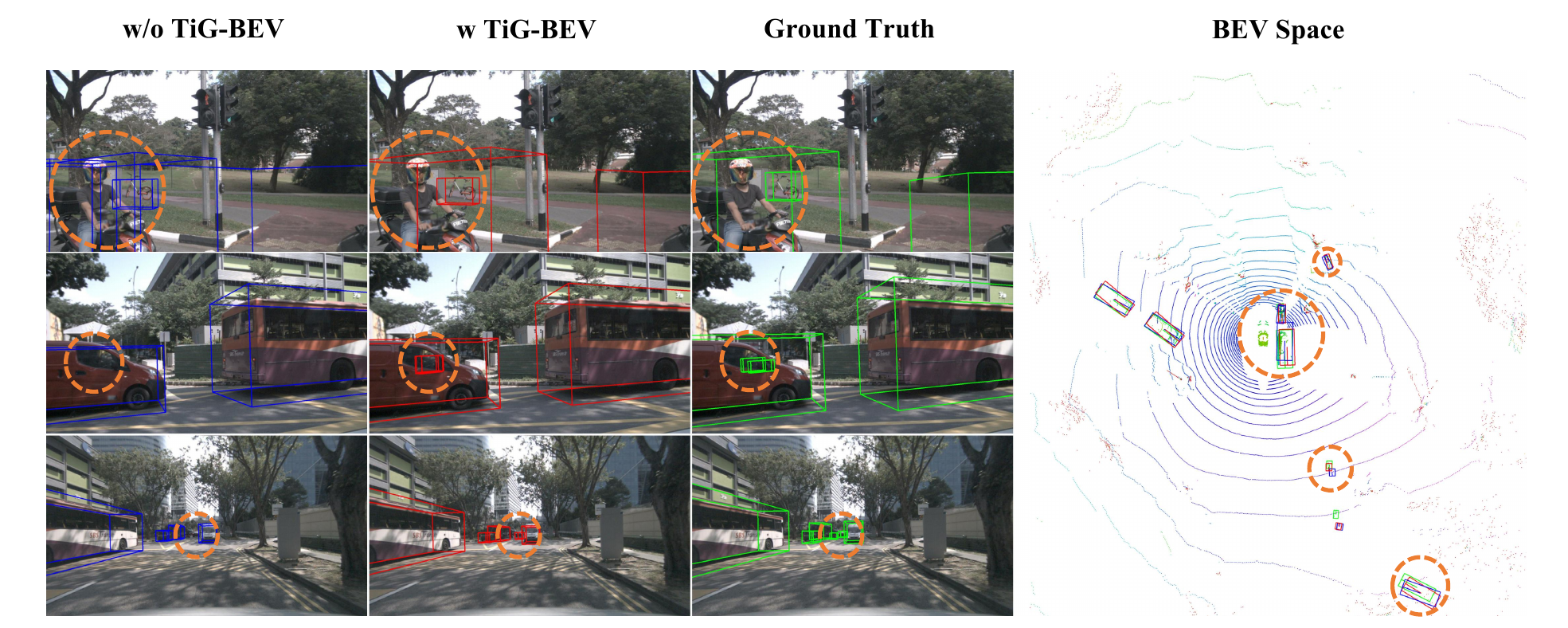}
    \caption{\textbf{Visualization of Detection Results}. From left to right, we show the 3D object detection before and after the TiG-BEV learning schemes, ground-truth annotations, along with the overall BEV-space results.
    }
    \label{fig:aabb}
\end{figure*}
\paragraph{Implementation Details.}
We implement our TiG-BEV using the BEVDet\cite{b19,b23} code base on 8 NVIDIA A100 GPUs, which is built on MMDetection3D toolkit\cite{b55}. A pre-trained CenterPoint~\cite{b53} with voxel size of $[0.1,0.1,0.2]$ is adopted as the LiDAR-based teacher, and the camera-based students include BEVDepth~\cite{b7}, BEVDet~\cite{b19} and BEVDet4D~\cite{b23}. During inference, the camare-based detecots only take multi-view images as input without the LiDAR data or teachers.
Referring to BEVDepth, we additionally add the dense depth supervision on top of BEVDet and BEVDet4D besides our TiG-BEV. We follow their official training configurations and hyperparameters as default, including data augmentation (random flip, scale and rotation), training schedule (2x), and others (AdamW optimizer~\cite{b16}, 2e-4 learning rate and batch size 8). For the main results on nuScenes val set in Table~\ref{tab:nus_val_sota}, we compare our TiG-BEV with all other methods under the CBGS strategy~\cite{b56}. For all other results, we do not utilize CBGS to better reveal the significance of proposed methods.




\subsection{Main Results}
\paragraph{On nuScenes Val Set.} 
In Table~\ref{tab:nus_val_sota}, we compare our TiG-BEV with other 3D object detectors on nuScenes val set. 
As shown, our LiDAR-to-camera learning schemes respectively boost the three baseline models, BEVDet, BEVDet4D, and BEVDepth, by +3.2\%, +2.6\%, and +2.3\% NDS. This clearly demonstrates the significance of our TiG-BEV to improve the detection performance of multi-view BEV 3D object detection.

\begin{table}[t!]
    \centering
    \caption{\textbf{Ablation Study of Target Inner-geometry Learning. }$\mathcal{L}^R_{\rm{depth}}$ and $\mathcal{L}_{\rm{bev}}$  denote the losses of inner-depth supervision and inner-feature BEV distillation, respectively.}
    \resizebox{1\columnwidth}{!}
    {\tablestyle{15pt}{1.0}
    \begin{tabular}{cc|cc}
    \toprule[1.2pt]
          $\mathcal{L}^R_{\rm{depth}}$&$\mathcal{L}_{\rm{bev}}$& mAP&NDS  \\
          \midrule
                        &           & 0.329         & 0.431 \\ 
             \checkmark &           & 0.339         & 0.440 \\
                        &\checkmark & 0.359         & 0.454 \\
              \checkmark&\checkmark & \textbf{0.366}         & \textbf{0.461} \\
    \bottomrule[1.2pt]
    \end{tabular}}
    \label{tab:main_ablation}
\end{table}

\paragraph{Without CBGS~\cite{b56} Strategy.} 
In Table~\ref{tab:aaaa}, we present the results of TiG-BEV without the CBGS training strategy. Without the resampling of training data, the performance improvement of learning target inner-geometry becomes more notable, \textbf{+10.3\%, +7.3\%,} and \textbf{+3.7\%} mAP for the three baselines, which indicates the superior LiDAR-to-camera knowledge transfer of our TiG-BEV.

\paragraph{Comparison with BEVDistill~\cite{b9}.} 
In Table~\ref{cccc}, we compare our TiG-BEV with another LiDAR-to-camera learning method BEVDistill in the same setting. As shown, on top of a better baseline model, our approach can achieve higher performance boost for both mAP and NDS. This well demonstrates the superiority of target inner-geometry learning to BEVDistill's foreground-guided dense distillation.



\begin{table}[t!]
    \centering
    \caption{\textbf{Ablation Study of Inner-depth Supervision.} We compare different settings for relative depth value calculation and depth reference selection. * denotes our implementation.}
    \resizebox{1\columnwidth}{!}
    {\tablestyle{7pt}{1.0}
    \begin{tabular}{c|c|cc}
    \toprule[1.2pt]
          Setting &Depth Reference & mAP&NDS  \\
          \midrule
          BEVDepth$^*$ & - & 0.329 & 0.431 \\
          \midrule
        \multirow{2}{*}{All-to-Certain}& 3D Center  & 0.358         & 0.452 \\ 
             &2D Center  & 0.358        & 0.452 \\
             \midrule
             One-to-One &Each Pair & 0.360     & 0.458 \\
             \midrule
         \multirow{2}{*}{All-to-Adaptive}&Highest Conf & 0.357         & 0.455 \\
              
              &Smallest Error & \textbf{0.366}         & \textbf{0.461} \\
    \bottomrule[1.2pt]
    \end{tabular}}
    \label{tab:relative_ablation}
\end{table}
\input{cvpr_2022/tex/tables/diffsetup_ablation}

\paragraph{Visualization.} 
As visualized in Figure~\ref{fig:aabb}, we show the detection results of BEVDepth before and after TiG-BEV, and the ground-truth annotations. We can clearly observe that more accurate results are obtained by our inner-geometry learning. Specifically, within the orange circles, the detection of false positives and ghosting objects can be reduced, and some 3D locations and orientations of the bounding boxes are also refined.



\subsection{Ablation Study}
\label{sec:ablation}
Here, we provide detailed experiments to validate the effectiveness of our approach from each of its components. We adopt BEVDepth as the student model by default.

\begin{figure}[t]
    \centering
    \includegraphics[scale=0.4]{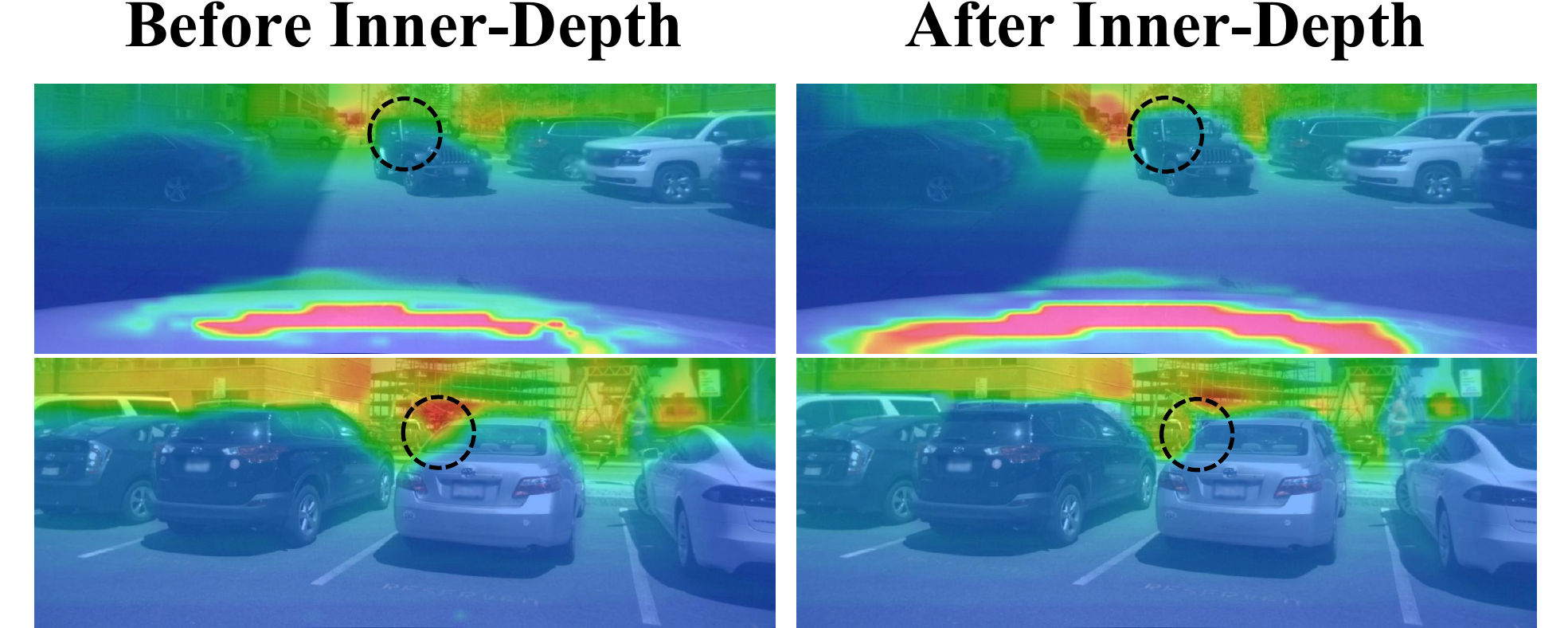}
    \caption{\textbf{Visualization of Predicted Depth Maps,} which are before and after the inner-depth supervision, respectively.}
    \label{fig:ref_compare}
\end{figure}

\paragraph{Inner-geometry Learning.} 
The individual effectiveness of the two main components can be examined by only equipping one of them. We first study the impact of the inner-depth supervision. As shown in Table~\ref{tab:main_ablation}, we only introduce the inner-depth supervision to the vanilla baseline, i.e., BEVDepth, by which its mAP improves from 32.9\% to 33.9\% with a +1.0\% gain and its NDS reaches 44.0\% from 43.1\% with a +0.9\% gain. 
Instead, when we only use the inner-feature BEV distillation, the mAP boosts from 32.9\% to 35.9\% with a +3.0\% improvement and the NDS boosts from 43.1\% to 45.4\% with a +2.3\% improvement. In addition, the combination of both components achieves better +3.7\% mAP and +2.7\% NDS, demonstrating that the two proposed objectives might be complementary.

\begin{table}[t!]
    \centering
    \caption{\textbf{Ablation Study of Inner-feature BEV Distillation.} $\mathcal{L}^{IC}_{\rm{bev}}$ and $\mathcal{L}^{IK}_{\rm{bev}}$ denote the losses of inter-channel and inter-keypoint distillation, respectively.}
    \resizebox{1\columnwidth}{!}
    {\tablestyle{15pt}{1.0}
    \begin{tabular}{cc|cc}
    \toprule[1.2pt]
          $\mathcal{L}^{IC}_{\rm{bev}}$&$\mathcal{L}^{IK}_{\rm{bev}}$& mAP&NDS  \\
          \midrule
                        &           & 0.329         & 0.431 \\ 
             \checkmark &           & 0.342         & 0.444 \\
                        &\checkmark & 0.358         & 0.452 \\
              \checkmark&\checkmark & \textbf{0.366}         & \textbf{0.461} \\
    \bottomrule[1.2pt]
    \end{tabular}}
    \label{tab:saf_ablation}
\end{table}

\paragraph{Inner-depth Supervision.}
To calculate the relative depth values within foreground targets, we compare several methods concerning the relationships among different inner points, which can be divided into three paradigms, 1) \emph{All-to-Certain} calculates the relative depth from all sampled points to a certain reference point, such as the projected center of 3D bounding box or the center of 2D bounding box. 2) \emph{All-to-Adaptive} sets the reference depth in a dynamic manner, which selects the reference pixel with the highest confidence across all depth bins or the smallest depth error to the ground truth (Ours). 3) \emph{One-to-One} calculates the relative depth from each two sampled point pair. As shown in Table~\ref{tab:relative_ablation}, compared with other patterns, our \emph{All-to-Adaptive} with smallest depth errors obtains the best performance improvement, which indicates the dynamic depth reference point can flexibly adapt to different targets for inner-geometry learning. What's more, we visualize our depth prediction with and without the inner-depth supervision in Figure~\ref{fig:ref_compare}, which effectively refines the contours and edges of foreground objects.

\paragraph{Inner-feature BEV Distillation.}
Our TiG-BEV explores the BEV feature distillation from two perspectives, inter-channel and inter-keypoint. To validate their effectiveness, we also equip the model with one of them at a time and report the results in Table~\ref{tab:saf_ablation}. As shown, both inter-channel and inter-keypoint distillation contribute to the final performance, respectively boosting the NDS by +1.3\% and +2.1\%. This well illustrates the importance of learning inner-geometry semantics within different foreground targets in BEV space. Further combining them two can benefit the performance by +3.7\% and +3.0\% for mAP and NDS.

\paragraph{2D Backbones and Temporal Information.}
We further explore the influence of 2D backbones and temporal information to our TiG-BEV in Table~\ref{tab:many_ablation}. We observe that our TiG-BEV brings significant performance improvement consistent over different 2D backbones. Also, our target inner-geometry learning schemes can provide positive effect for both single-frame and multi-frame settings. The improvement of mAP ranges from \textbf{+3.4\%} to \textbf{+5.8\%} and the improvement of NDS ranges from \textbf{+2.5\%} to \textbf{+5.0\%}.

%% file: cvpr_2022/tex/tables/diffsetup_ablation.tex
\begin{table*}[ht]
\centering
\caption{\textbf{Ablation Study of 2D Backbones and Temporal Information.} CenterPoint~\cite{b53} and BEVDepth~\cite{b7} are adopted as the teacher and student models, respectively.}
\resizebox{2\columnwidth}{!}
{\tablestyle{12pt}{1}
\begin{tabular}{c|c|c|c|cc}
\toprule[1.2pt]
          Backbone&Resolution&Multi-frame&Method& mAP&NDS  \\
          \midrule
          VoxelNet & - & \checkmark&Teacher & 0.564 & 0.646 \\\midrule
          \multirow{4}{*}{ResNet-18}&\multirow{4}{*}{$256\times 704$}&\multirow{2}{*}{\checkmark}&Student & 0.285 & 0.405 \\ 
              & & &  + TiG-BEV& \textbf{0.323 (\textcolor{blue}{+3.8\%})}         & \textbf{0.430 (\textcolor{blue}{+2.5\%})} \\
              \cmidrule(lr){3-6}
                & & &  Student& 0.260  & 0.295 \\
               & & &    + TiG-BEV& \textbf{0.294 (\textcolor{blue}{+3.4\%})}  & \textbf{0.335 (\textcolor{blue}{+4.0\%})}\\
          \midrule
         \multirow{4}{*}{ResNet-50}&\multirow{4}{*}{$256\times 704$}&\multirow{2}{*}{\checkmark}&Student & 0.329 & 0.431 \\ 
              & & &    + TiG-BEV& \textbf{0.366 (\textcolor{blue}{+3.7\%})}         & \textbf{0.461 (\textcolor{blue}{+3.0\%})} \\
              \cmidrule(lr){3-6}
              
              & & &  Student& 0.298  & 0.328 \\
               & & &  + TiG-BEV& \textbf{0.338 (\textcolor{blue}{+4.0\%})}  & \textbf{0.375 (\textcolor{blue}{+4.7\%})}\\
          \midrule
          \multirow{4}{*}{ResNet-101}&\multirow{4}{*}{$512\times 1408$}&\multirow{2}{*}{\checkmark}&Student & 0.393 & 0.487 \\ 
              & & &  + TiG-BEV& \textbf{0.430 (\textcolor{blue}{+3.7\%})}         & \textbf{0.514 (\textcolor{blue}{+2.7\%})} \\
              \cmidrule(lr){3-6}
              & & &  Student& 0.345 & 0.366 \\
               & & &    + TiG-BEV& \textbf{0.403 (\textcolor{blue}{+5.8\%})}        & \textbf{0.416 (\textcolor{blue}{+5.0\%)}} 
         \\
               \bottomrule[1.2pt]
    \end{tabular}}
    \label{tab:many_ablation}
\end{table*}